\documentclass[runningheads]{llncs}

\usepackage{eccv}

\usepackage{eccvabbrv}
\usepackage{graphicx}
\usepackage{booktabs}
\usepackage[accsupp]{axessibility}  

\usepackage{booktabs}
\usepackage{mathtools}
\usepackage{multirow}
\usepackage{bm}
\usepackage{algorithm}
\usepackage{algpseudocode}
\usepackage{pifont}

\usepackage{graphicx}
\usepackage{amsmath}
\usepackage{amssymb}
\usepackage{wrapfig,lipsum}
\usepackage{svg}

\usepackage[pagebackref,breaklinks,colorlinks,citecolor=eccvblue]{hyperref}

\usepackage{orcidlink}
\newcommand{\notcomp}[1]{{\color{gray}#1}}

\newcommand{\sst}[1]{{\color{black}#1}}

\newcommand{\sg}[1]{{\color{black}#1}}

\begin{document}

\title{Training-free Zero-shot Composed Image Retrieval with Local Concept \sg{Re-}ranking} 

\titlerunning{Abbreviated paper title}

\author{Shitong Sun\inst{1} \and
Fanghua Ye\inst{2}\and
Shaogang Gong\inst{1}}


\institute{Queen Mary University of London\and
University College London
}

\maketitle

\newcommand{\method}{TFCIR\xspace}
\newcommand{\rerank}{LCR\xspace}
\newcommand{\baseline}{GRB\xspace}

\begin{abstract}
  Composed image retrieval attempts to retrieve an image of interest from gallery images through a composed query of a reference image and its corresponding modified text. 
  It has recently attracted attention due to the collaboration of information-rich images and concise language to precisely express the requirements of target images. 
  Most \sg{current} composed image retrieval methods follow a
  supervised learning \sg{approach} to training on a costly triplet
  dataset composed of a reference image, modified text, and a
  corresponding target image. 
  To \sg{avoid} difficult-to-obtain labeled triplet \sg{training data,
    zero-shot composed image retrieval (ZS-CIR) has been introduced}, which
  aims to retrieve the target image \sg{by learning from image-text
    pairs (self-supervised triplets),} without the need for human-labeled triplets. 
  However, \sg{this self-supervised triplet learning approach is
    computationally less effective and less understandable as it
    assumes} the interaction between image and text is conducted with
    implicit query embedding \sg{without explicit semantical interpretation}. 
  In this work, we present a new training-free zero-shot composed image retrieval method which translates the query into explicit human-understandable text.
  This helps improve \sg{model learning efficiency to enhance the generalization capacity} of foundation models.
  Further, we introduce a Local Concept \sg{Re-}ranking (\rerank) mechanism to focus on discriminative local information extracted from the modified instruction\sg{s}. 
  %
  Extensive experiments on four ZS-CIR benchmarks show that \sg{our
    method achieves} comparable performances \sg{to that of the} state-of-the-art
  \sg{triplet training based} methods
  , \sg{but} significantly \sg{outperforms other training-free} methods on
  the open domain datasets \sg{(CIRR, CIRCO and COCO)}, as well as the
  fashion domain dataset \sg{(FashionIQ)}. 
   
  \keywords{Multimodal \and Image retrieval \and Training free }
\end{abstract}

\begin{figure}[t]
  \centering
  \includegraphics[width=0.8\textwidth]{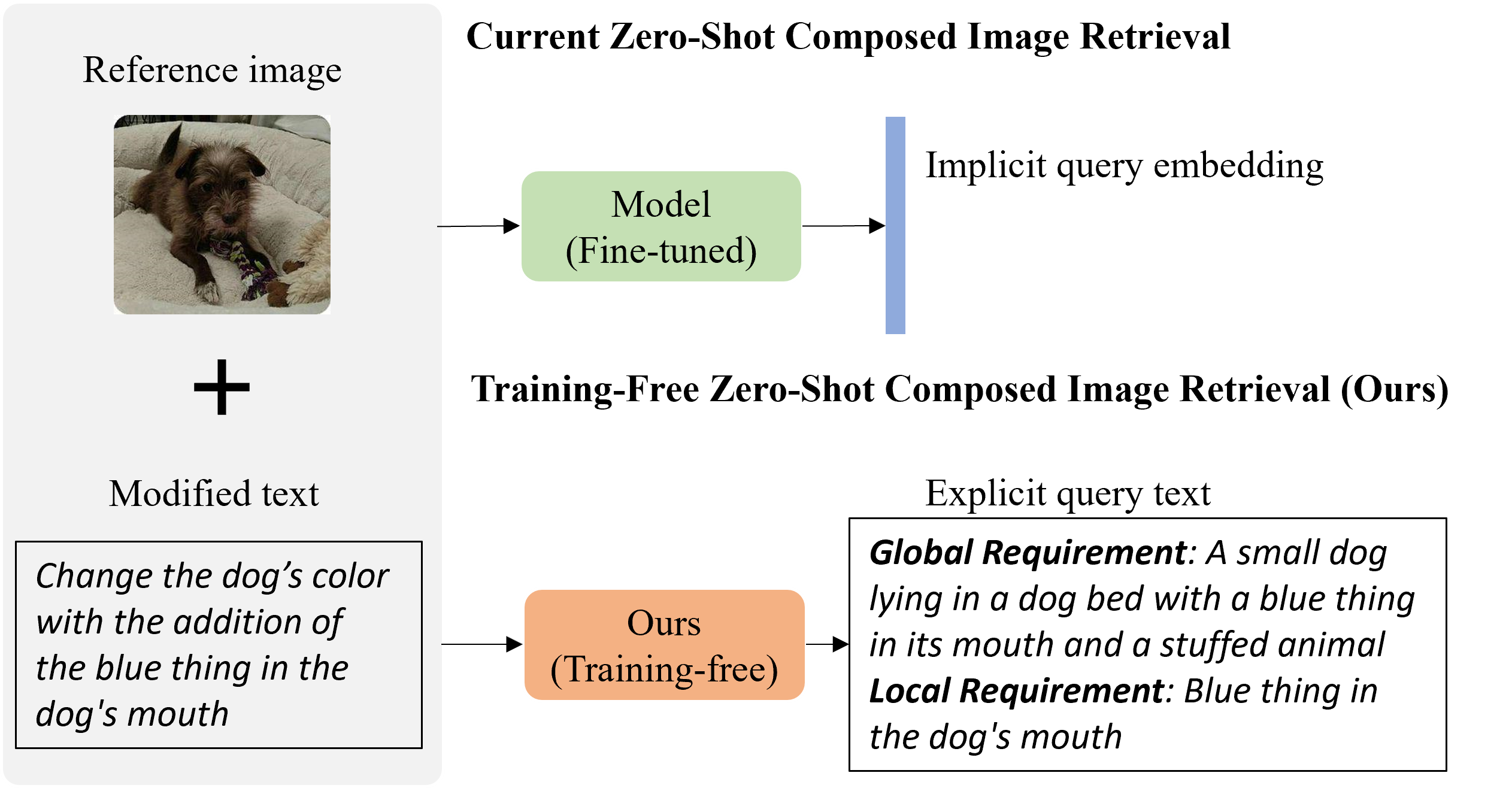}
  \caption{A comparison of query processing between existing zero-shot
    composed image retrieval methods and \sg{our method}. 
  \textbf{Top:} Existing methods are fine-tuned on text-image pairs~\cite{saito2023pic2word,baldrati2023zero} or task-specific triplets~\cite{levy2023data,liu2023zero}.
 \sg{They} employ a fusion embedding for queries lacking explicit semantic context. 
  \textbf{Bottom:} We propose a novel method without the requirement of training. 
  Our method processes queries with explicit semantics at both global and local levels.
  } \vspace{-0.5cm}
  \label{fig:intro}
\end{figure}

\section{Introduction}
\label{sec:intro}
Composed image retrieval (CIR) aims to retrieve the target image from gallery images through the composed query, which is specified in the form of an image plus some text that describes desired modifications to the input image~\cite{vo2019composing,chen2020image}. 
%
%
The text-image composed query can overcome the limitation of singular modality image retrieval, where the image is dense in semantics and text is sparse~\cite{gabbay2021image}.
Specifically, text-image retrieval encounters challenges due to imprecise descriptions and an unlimited number of correct targets. For example, a text query `dog' can correspond to thousands of dog images in various breeds, poses, and environments.
Meanwhile, image-image retrieval suffers from an expression limitation, lacking the ability to generalize to different visual contents~\cite{gordo2016deep}. 
This becomes infeasible when a single image cannot be tailored to users' intentions.

Traditional composed image retrieval methods~\cite{liu2021image,vo2019composing,chen2020image,baldrati2022effective,delmas2022artemis} are primarily trained on datasets of triplets composed of a reference image, a relative caption, and a target image. 
However, these methods entail the expensive cost of collecting triplet datasets.
To overcome this challenge, recent research has made efforts to consider zero-shot composed image retrieval (ZS-CIR).
Specifically, current methods for ZS-CIR are mainly based on the nearest neighbor search method in pre-trained visual-language embedding spaces~\cite{radford2021learning,li2022blip}.
These embeddings are further fine-tuned using either task-independent image-text pairs~\cite{saito2023pic2word,baldrati2023zero} or additional task-specific image-text-image triplets~\cite{levy2023data,liu2023zero}.
%

However, this direct usage \sg{approach} has two \sg{drawbacks:}
Firstly, 
a large amount of computational resources
is required for training large
foundation models on downstream image-text
pairs~\cite{saito2023pic2word,baldrati2023zero} or task-specific
triplets~\cite{levy2023data,liu2023zero} which reduces model learning efficiency;
Secondly, current methods are less \sg{transparent to human understanding}. 
They represent \sg{a} composed query as an implicit holistic
embedding, assuming that the interaction between the reference image
and modified text is \sg{implicitly} encoded. 
However, typical interactions that encode image-image relations such
as addition, negation, comparison, spatial relations
\sg{variation}~\cite{baldrati2023zero} are not learned in the
pretraining processes of foundation
models~\cite{radford2021learning,li2022blip}, which primarily focus
training with image-text pairs. \sg{Therefore, this implicit encoding
  assumption is computationally {\em not} enforced nor necessarily valid.}
Specifically, in image-text pairs, the text is a descriptive image caption, aligning with the visual content of the target image. In contrast, in image-text-image triplets, the text is an instruction encoding variations of the reference visual content, which can be irrelevant to the target image. 
Using text-image aligned space~\cite{radford2021learning} directly,
\sg{such as}~\cite{saito2023pic2word,baldrati2023zero}, to
\sg{represent change instruction is sub-optimal}.

To address the aforementioned problem, we have taken the lead to build a \textit{training-free} baseline with explicit human-understandable query  
for zero-shot composed image retrieval.
The comparison of query processing between conventional methods and our proposed methods is shown in Figure~\ref{fig:intro}.
Our method is designed to convert the composed query, consisting of a reference image and modified text, into a single text query. 
It retrieves the target image in \sg{an} aligned vision-language
feature space~\cite{li2023blip} without the requirement for additional
training. 
Utilizing the recently developed captioner BLIP2~\cite{li2023blip} to translate the image to text modality and a large language model for text understanding, our mechanism is not only computationally efficient, being training-free, but also capable of generating human-understandable query text. 
%
%
\sst{Nevertheless, querying with a single holistic query text can be misleading,
which may include distractions from the rich semantic information encoded in the reference image.}
%
Specifically, for the composed query, the modified text presents a hard precise requirement, while the reference image provides a soft ambiguous requirement.
As the reference image encodes rich visual concepts, but it is unknown whether to preserve visual concepts that are not mentioned in the modified text.
%
To address this challenge, we refine the hard requirement, which represents modification of the reference image, into local concepts that are required to exist in the target image.
Furthermore, we design a re-ranking mechanism based on these hard local concepts to provide precise and discriminative query requirements. 
%

Our work is motivated by the recent development of 
vision-language models~\cite{liu2023visual}, which excel in local feature discrimination. 
However, properly converting implicit text-image queries to explicit text-only queries at the local level remains a challenge. 
The vision-language model~\cite{liu2023visual} is designed to generate text but lacks the ability to rank multiple gallery images.
To address this limitation, we introduce a scoring module based on the probability of text prediction.
Specifically, we extract local discriminative text as a text understanding problem and force the vision-language model to judge whether the local attributes exist in the target images by answering `Yes' or `No'. 
Together with global ranking score 
of the query text and target images, we incorporate the probability of text prediction as re-ranking scores. 
With this local concept re-ranking mechanism, our model can generate human-understandable explicit attributes in the training-free framework.
\sg{Our contributions are: (1) We introduce for the first time} a
training-free method for zero-shot composed image retrieval,
eliminating the need for training on either image-text pairs or
task-specific triplets; \sg{(2)} We introduce a two-stage approach
which incorporates coarse global-level retrieval (\baseline) as the
first stage and fine-grained local concept re-ranking (\rerank) as the
second stage.  
The first stage \baseline tunes the instruction-image pair into a descriptive caption to leverage existing visual-textual alignment space. 
The second stage \rerank solves the problem of ambiguous
requirement from reference image by re-ranking on precise fine-grained
concepts.  
  This multiple-criteria scoring mechanism enables the model to rank based on both global and local information;
  \sg{(3)} Our method achieves comparable performances \sg{to that of the}
  state-of-the-art \sg{training based} methods \sg{but outperforms
    significantly all other} training-free methods \sg{for open domain test}
    on four different datasets: CIRR, FashionIQ, CIRCO and COCO. 
\begin{figure*}[t]
  \centering
  \includegraphics[width=1.0\textwidth]{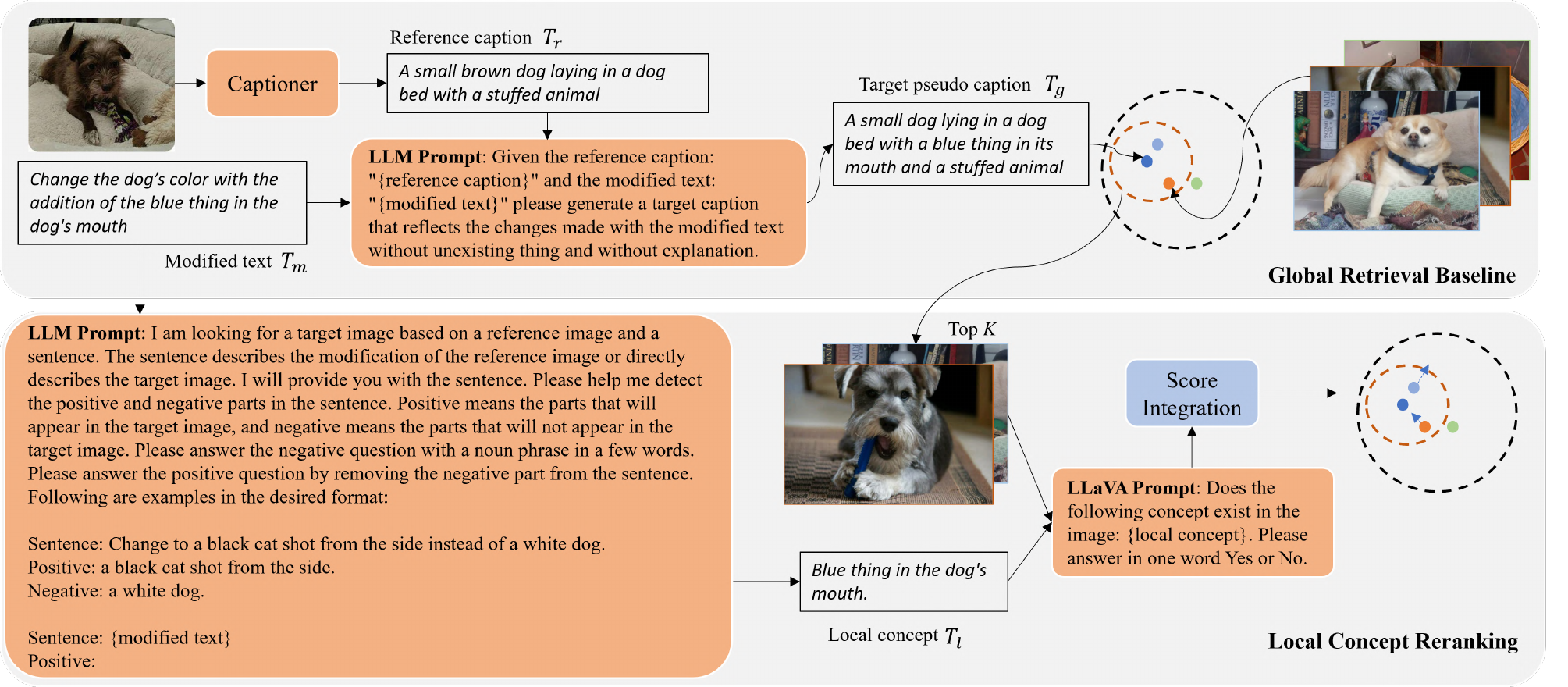}
  \caption{An overview of the Training-Free Composed Image Retrieval model which consists of \textbf{Top:} Global Retrieval Baseline (GRB) transforms the text-image composed query into a text-only query with a Large Language Model (LLM)-generated pseudo target caption. 
  The global score for retrieval is determined by the similarity between the pseudo target caption's text embedding and target image embeddings.
  \textbf{Bottom:} Local Concept Re-Ranking (LCR) extracts discriminative local concepts using a task instruction prompt for the LLM. 
  Based on the global score from GRB, the top-$K$ images are re-ranked using both the global and local scores, calculated on their discriminative local concepts.
  The local score is translated into text prediction probability, detecting the existence of local concepts by a vision-language model, i.e. LLaVA.  
  Both the global and local retrieval processes are conducted within a visual-language aligned feature space, i.e. BLIP2.} \vspace{-0.3cm}
  \label{fig:overview}
\end{figure*}

\section{Related \sg{Works}}
\label{sec:related_work}
\textbf{Composed Image Retrieval.}
Composed image retrieval aims to retrieve the target image from a gallery set where the input query is in the form of more than one modality. 
A standard query is composed of an image and various interactions, including relative attributes~\cite{parikh2011relative}, natural language~\cite{chen2020image, vo2019composing} or spatial layout~\cite{mai2017spatial}, to describe the desired modifications.
Among these modalities, natural language has gained attention as it serves as a common means of conveying intricate specifications between humans and computers.
Consequently, composed image retrieval becomes almost indistinguishable from image retrieval queried by both image and modified text. 
The key challenge in this task lies in learning the 
relation between images. 
Conventional composed image retrieval 
addresses this challenge by training on task-specific triplets datasets. 
However, the demand for triplets datasets in CIR has led to expensive costs.
To alleviate the need for triplets data, recent methods 
explore zero-shot composed image retrieval without training on existing benchmarks. 
Among them, CASE~\cite{levy2023data} and TransAgg~\cite{liu2023zero} construct additional task-specific triplets datasets, which lack potential scalability.
Pic2Word~\cite{saito2023pic2word} and SEARLE~\cite{baldrati2023zero} avoid using task-specific triplets but perform fine-tuning on text-image pairs by converting reference images into a single token. 
All existing zero-shot composed image retrieval methods require a training stage and compromise the generalization of foundation models. 
We \sg{introduce for the first time} a training-free method for zero-shot composed image retrieval.



\noindent\textbf{Descriptive LLMs in Vision.}
The advancements in foundation models open up new opportunities for understandable visual recognition. 
Recent works~\cite{yan2023learning,yang2023language,menon2022visual} have mainly focused on image classification tasks, generating class-level attributes from class labels using the prior knowledge of the language model. 
The key idea is to query the large language model to obtain a set of attributes that describe each class and further apply vision-language models~\cite{radford2021learning} to classify images via LLM-generated attributes.
These fine-grained attribute-based classifications inherently provide interpretability.
However, all of the current works are based on class level and cannot be generalized to the broader topic of instance level. 
In this work, we extend the descriptive LLMs to generate instance-level attributes in a composed image retrieval task.
%
Another relevant work in composed image retrieval CASE~\cite{levy2023data} provides the interpretability through visualization of gradient-based heatmaps on both image and text. 
However, this visualization can only know where the model looks instead of what the model sees.
In contrast, we aim to provide concept-level attributes instead of pixel-level saliency.


\noindent\textbf{Prompt Engineering.}
With the rapid development of large foundation models and the increase of training difficulty, the paradigm has gradually transferred from traditional supervised learning to more efficient prompt-based learning~\cite{liu2023pre}. This approach involves applying various templates to model input and leveraging zero-shot or few-shot learning to transfer model knowledge to various downstream scenarios.  
This prompt engineering topic originates from the natural language processing domain~\cite{brown2020language,radford2019language} and is later introduced to the multimodality domain~\cite{zhou2022conditional,zhou2022learning}. 
Extensive prompts~\cite{gu2023systematic} have been explored including task instruction prompting~\cite{radford2019language,efrat2020turking} for downstream tasks, in-context learning~\cite{dong2022survey,brown2020language} by showing examplars, chain-of-thought prompting~\cite{press2022measuring,wei2022chain} for progressively leading to easier logic. 
In this work, we explore the task instruction prompting for composed image retrieval. 
Specifically, leveraging the power of language models, we explore one prompt for merging reference image captions and another prompt for extracting local concepts with randomly selected examples.
This prompt engineering is designed to stimulate the reasoning ability of large language models. 


\section{Methodology}
\label{sec:methodology}
Given a reference image $I_r$ and its corresponding modified text $T_m$ as input query, the aim of composed image retrieval is to identify the target image $I_t$ from the gallery set $\{I_t^n\}_{n=1}^N$ - so that the target image accurately reflects the interaction between the reference image and modification described in the text.
It is challenging to deal with the interaction between two fundamentally different modalities: images encode dense semantic information within a continuous visual space, while modified text encodes sparse semantic information in a discrete textual space. 
Meanwhile, reference image represents an ambiguous requirement where visual contents can change, while modified text represents a precise requirement where all mentioned interactions need to be satisfied. 

In this work, we study the training-free zero-shot composed image retrieval problem which confronts the challenges of both inherent modality gap and ambiguous soft requirements provided by the reference image. 
To that end, as shown in Figure~\ref{fig:overview}, we propose a multi-level ranking mechanism with a coarse global retrieval considering both soft and hard requirements for the first stage (Section~\ref{sec:global}) and a local concept re-ranking focusing on hard requirement as the second stage (Section~\ref{sec:local}).
We leverage vision-language models to bridge the modality gap between image and text, and further utilise language models to deal with the modification reasoning.
The aim of the proposed method is to preserve the generalization ability of the foundation model by being training-free, as well as enabling retrieval with both global features about holistic query information and local features about discriminative partial query information.

%
%
\subsection{Global Retrieval Baseline}
\label{sec:global}

To retrieve images of interest from composed image-text queries in a training-free paradigm, we formulate a Global Retrieval Baseline (\baseline) which converts the less-explored composed image retrieval task into the well-developed text-image retrieval task.
Namely, we convert the text-image composed query into a text-only query and use existing aligned visual-language feature space~\cite{li2023blip} to retrieve the target image of interest.  
As shown in the upper part of Figure~\ref{fig:overview}, the overall pipeline of our global retrieval baseline consists of three steps, including reference image captioning, pseudo target caption generation, and text-image retrieval through the aligned feature space. 

To bridge the modality gap between image and text, we leverage off-the-shelf captioner BLIP2~\cite{li2023blip} to generate a brief caption $T_r$ of the reference image $I_r$ defined as:
\begin{equation}
  \label{eq:reference_caption}
  T_r = \text{Captioner} (I_r).    
\end{equation}

Instead of dealing with a multimodal query fusion problem, we utilise \sg{a training-free text-image space} for image retrieval.
To this end, \sg{we construct} a single text query based on our
generated caption $T_r$ and modified text $T_m$, which is a
language-only task with a requirement of instruction reasoning. 
Notably, conventional text-image aligned space~\cite{radford2021learning, li2023blip} aligns an image and its corresponding caption, a description of the image's visual content.
However, in composed image retrieval, the relation between the
reference and target images is encoded in the modified text, an
instruction for adjusting the image including addition, negation, \sg{and comparison. }
In other words, every word in conventional text-image retrieval has its counterpart in the corresponding image and contributes to boosting the retrieval. 
In contrast, in composed image retrieval, not all words have a direct correspondence with the target image, particularly for negation of the original image. 
For example, \textit{`Change the dog to a cat'} or \textit{`Remove the lemon'} illustrate these scenarios, where \textit{`the dog'} and \textit{`the lemon'} should not appear in the target image.
In light of this, directly combining the generated reference image caption $T_r$ and the modified text $T_m$ can lead to misalignment.
%
Therefore, we leverage the reasoning ability of Large Language Model (LLM), GPT4 in particular, to generate a pseudo target caption $T_g$ as the global-level retrieval information. We have:
\begin{equation}
  \label{eq:global_caption}
  T_g = \text{LLM} (P_g, T_r, T_m).
\end{equation}
The prompt $P_g$ to generate a global-level pseudo target caption is shown in Figure~\ref{fig:overview}.

%
Using the pseudo target caption, the global retrieval ranks target images by similarity of the text embedding of pseudo target caption $T_g$ and image embeddings of target images in the pre-learned text-image aligned feature space~\cite{li2023blip}. The global score for target image $I^n_t$ is calculated as follows:
\begin{equation}
    \label{eq:global_retrieval}
    S_{g}^n = <E_{t}(T_g),E_{i}(I_t^n)>,
\end{equation}
where $< >$ is to calculate the cosine similarity, $E_{t}$ and $E_{i}$ are the text and image encoders of BLIP2~\cite{li2023blip} respectively. 
For each query in this inference process, Eq.~\eqref{eq:reference_caption}-\eqref{eq:global_caption} are conducted once and Eq.~\eqref{eq:global_retrieval} is conducted $N$ times for all the target images. 
Consequently, the multimodal fusing problem with the inherent modality gap has been converted into a text comprehension problem, and the visual-textual aligned space can be leveraged directly without training.


\subsection{Local Concept Re-Ranking}
\label{sec:local} 
Even though the formulated baseline with an explicit semantic query is able to transfer a text-image composed query to a text-only query in a training-free method,
%
it still suffers from the ambiguous requirement provided by the reference image. 
As an image encodes \sst{rich} semantic information, a directly generated image
caption can include visual concepts not expected in the target image,
which will distract the retrieval process.
%
%
%
In light of this, identifying the local visual concepts that will actually appear in the target image instead of all visual concepts from the generated query caption $T_g$, can provide discriminative information and reduce false positives. 
Therefore, we introduce a Local Concept Re-Ranking (\rerank) module to extract discriminative concepts that must exist in target images to rerank the results from the global retrieval baseline.
As depicted in the lower part of Figure~\ref{fig:overview}, the \rerank module consists of two steps: local concept re-identification and scoring integration.  
This local concept re-ranking mechanism, which identifies the concepts that must exist in the target image, enhances discriminative information from an ambiguous global level in the \baseline to a precise fine-grained local level. 
%



\subsubsection*{Local Concept Re-identification.}

To identify the local concepts that must exist in the target image, we parse the modified text $T_m$ by a large language model.
We refer to these local concepts that must exist in the target image as positive and those that should not exist as negative.
By designing a local concept parsing prompt $P_l$ with manually crafted exemplars as shown in Figure~\ref{fig:overview}, we obtain the positive concepts in textual form by removing the negative concepts from the modified text.
Notably, it becomes popular to leverage LLMs to generate text descriptions~\cite{yan2023learning,yang2023language,menon2022visual} to enhance visual perception. 
However, current descriptions are typically generated at the class level while our approach generates descriptions at the instance level, specifically tailored to each query.
The process can be formulated as follows:
\begin{equation}
  \label{eq:modified_parse}
  T_l = \text{LLM} (P_l, T_m),
\end{equation}
where $T_l$ denotes the positive local concepts in textual form, providing precise information to enhance discriminative ability.
%
%





Having obtained the positive concepts by parsing modified text in the query, we further determine whether the positive concepts exist for each target image candidate. 
To detect the local concept existence, we utilize a multimodal model LLaVA~\cite{liu2023visual}, which is an instruction-following visual question-answering model and returns the result by text. 
However, LLaVA can only detect the existence of local visual concepts and return a hard result with the text `Yes' or `No', which cannot be utilized in the ranking scenario directly. 
To address this issue, we unlock the last layer of the LLaVA.
Specifically during text generation, LLaVA predicts a score for each token and calculates the next token probability distribution. We focus on the score of tokens ‘Yes’ and ‘No’, and further calculate the visual concept existence probability by exponential normalization, which can be formulated as follows:
\begin{equation}
  \label{eq:reidentification}
  \text{Logit}_{\text{YES}}^n,\text{Logit}_{\text{NO}}^n = \text{LLaVA} (P_{re}, T_l, I^n_t),  
\end{equation}
\begin{equation}
  \label{eq:reidentification_soft}
  S_{l}^n = softmax(\text{Logit}_{\text{YES}}^n,\text{Logit}_{\text{NO}}^n),
\end{equation}
where $P_{re}$ is the prompt for LLaVA as shown in Figure~\ref{fig:overview} and $S_{l}^n$ is the local score for the $n$-th target image.
To enhance efficiency and reduce computation, we only re-rank the top-$K$ results of the global baseline.
This explicit comparison can help re-identify the fine-grained discriminative information and thus boost retrieval performance. 

\subsubsection*{Score Integration Module.}
Given the ranking scores from the first stage of the global retrieval baseline and the second stage of local concept re-ranking, we can achieve the final ranking score by simply summing the global score and local score.
Namely, the final score $S^n$ for the $n$-th target image can be shown as follows: 
\begin{equation}
  \label{eq:score_module}
  S^n = S_g^n + \gamma \cdot S_l^n.
\end{equation}
The final ranking is achieved by the $S^n$ value from maximum to minimum.  
The scale parameter $\gamma$ is a hyperparameter with a default value equal to one.  
We summarize our proposed method in Algorithm~\ref{algo:fzsl}.

\begin{algorithm}[t]

    \caption{\textbf{Training-Free Zero-Shot Composed Image Retrieval with Local Concept Re-Ranking}.
    }
  \label{algo:fzsl}
\begin{algorithmic}[1]
\State \textbf{Input:} Query of reference image $I_r$ and modified text $T_m$. Gallery set of target images \{$I_{t} \}_{n=1}^N$. Hyperparameter rerank top $K$ and scale parameter $\gamma$.

  \State \textbf{Global Retrieval Baseline:}  \\
 Generate reference image caption $T_r $ via Eq.~\eqref{eq:reference_caption}. \\
 Generate pseudo target caption at the global level $T_g$ via Eq.~\eqref{eq:global_caption}.  \\
 Calculate global score $S^n_g$ via Eq.~\eqref{eq:global_retrieval} and achieve top-$K$ ranking results from maximum to minimum.
\State \textbf{Local Concept Re-Ranking:}  \\
Modification parsing to extract local positive concepts $T_l$ via Eq.~\eqref{eq:modified_parse}. 
\For{each target image $n$ = 1 to $K$ in the top-$K$ ranking results returned by the global retrieval baseline}
\State {Calculate local score $S_l^n$ via Eq.~\eqref{eq:reidentification} and Eq.~\eqref{eq:reidentification_soft}}.
\EndFor \\
Calculate final score $S^n$ for target images via Eq.~\eqref{eq:score_module}.
 
\end{algorithmic}
\end{algorithm}

\section{Experiments}
\label{sec:experiment}
\subsection{Datasets and Evaluation Metrics}

We conduct experiments on three publicly available CIR benchmarks, CIRR~\cite{liu2021image} and CIRCO~\cite{baldrati2023zero} for scene manipulation, and FashionIQ~\cite{guo2019fashion} for fashion attribute manipulation. Additionally, we utilize a general dataset 
COCO~\cite{lin2014microsoft} for object composition following~\cite{saito2023pic2word}. 
The three CIR benchmarks are in the form of triplets, composed of a reference image, modified text, and one or multiple target images.
FashionIQ, focused on the fashion domain, contains 
garment images that can be divided into three categories: dress, shirt and toptee. 
%
%
CIRR is composed of 
real-life images extracted from NLVR2~\cite{suhr2018corpus}, which contains rich visual content against various backgrounds. 
CIRCO, an open-domain dataset, 
utilizes all 120K images of COCO~\cite{lin2014microsoft} as the gallery set and provides multiple ground truths for each query. 
For object composition following~\cite{saito2023pic2word}, cropped images from COCO validation set (5,000 images) with masked backgrounds are used as queries, with the modified text consisting of a list of object classes.
We adopt zero-shot evaluation to test on the test sets of CIRR and CIRCO, and the validation set of FashionIQ following the standard evaluation protocol~\cite{saito2023pic2word,baldrati2023zero}. 
%
%
%

\begin{table}[!t]
  \centering
  \caption{\textbf{Quantitative results on CIRR test set.} Best scores are in bold. } 
   \resizebox{1.0\linewidth}{!}{ 

   \begin{tabular}{llccccccc} 
    \toprule
    \multicolumn{1}{l}{} & \multicolumn{1}{c}{} & \multicolumn{4}{c}{Recall} & \multicolumn{3}{c}{Recall$_{\text{Subset}}$} \\
    \cmidrule(lr){3-6}
    \cmidrule(lr){7-9}

    \multicolumn{1}{l}{Supervision} & \multicolumn{1}{l}{Method} & $R@1$ & $R@5$ & $R@10$ & $R@50$ & $R@1$ & $R@2$ & $R@3$\\ 
   \midrule
   \multirow{3}{*}{Triplets }  & \textcolor{gray}{CASE}~\cite{levy2023data} & \textcolor{gray}{35.40} & \notcomp{65.78} & \notcomp{78.53} & \notcomp{94.63} &\notcomp{ 64.29} & \notcomp{82.66} & \notcomp{91.61} \\ 
  & \textcolor{gray}{TransAgg-combined}~\cite{liu2023zero}&  \notcomp{37.87} & \notcomp{68.88} & - & \notcomp{93.86}  & \notcomp{69.79} & - & -\\ 
  & \textcolor{gray}{TransAgg-template}~\cite{liu2023zero}&  \notcomp{38.10} & \notcomp{68.42} & - & \notcomp{93.51}  & \notcomp{70.34} & - & -\\ 
  \midrule[.02em]  
  \multirow{3}{*}{Text-image pairs} & PALAVRA~\cite{cohen2022my}  & 16.62 & 43.49 & 58.51 & 83.95 & 41.61 & 65.30 & 80.94\\

   & Pic2Word~\cite{saito2023pic2word} & 23.90 & 51.70 & 65.30 & 87.80  & - & - & - \\
   & SEARLE~\cite{baldrati2023zero} & 24.87 & 52.31 & 66.29 & \textbf{88.58} & 53.80 &74.31 &86.94 \\

 \midrule[.02em]

 \multirow{5}{*}{Training-free}& Text-only-M & 21.49 & 45.78 & 57.11 & 81.33 & 64.80 & \textbf{82.72} & \textbf{91.78}\\  
 & Text-only-C  & 13.66 & 37.78 & 51.01 &  80.60 & 42.65 & 66.77 & 82.87 \\  
 & LLaVA-CIR  & 11.90 & 29.21 & 39.74 &62.96  & 37.71 & 57.52 & 74.43\\  

 \cmidrule(lr){2-9}
& \baseline  & 24.19 & 52.07 & 64.48 & 85.28 & 60.17 & 79.13 & 89.34\\  

 & \baseline + \rerank & \textbf{32.82} & \textbf{61.13} &\textbf{ 71.76} & 85.28 & \textbf{66.63} & 78.58 & 82.68 \\  

\bottomrule
   \end{tabular}}
   \vspace{-0.5cm}
   \label{tab:cirr_test_exp}
 \end{table}

 \begin{figure*}[t]
	\centering
	\includegraphics[width=0.95\textwidth]{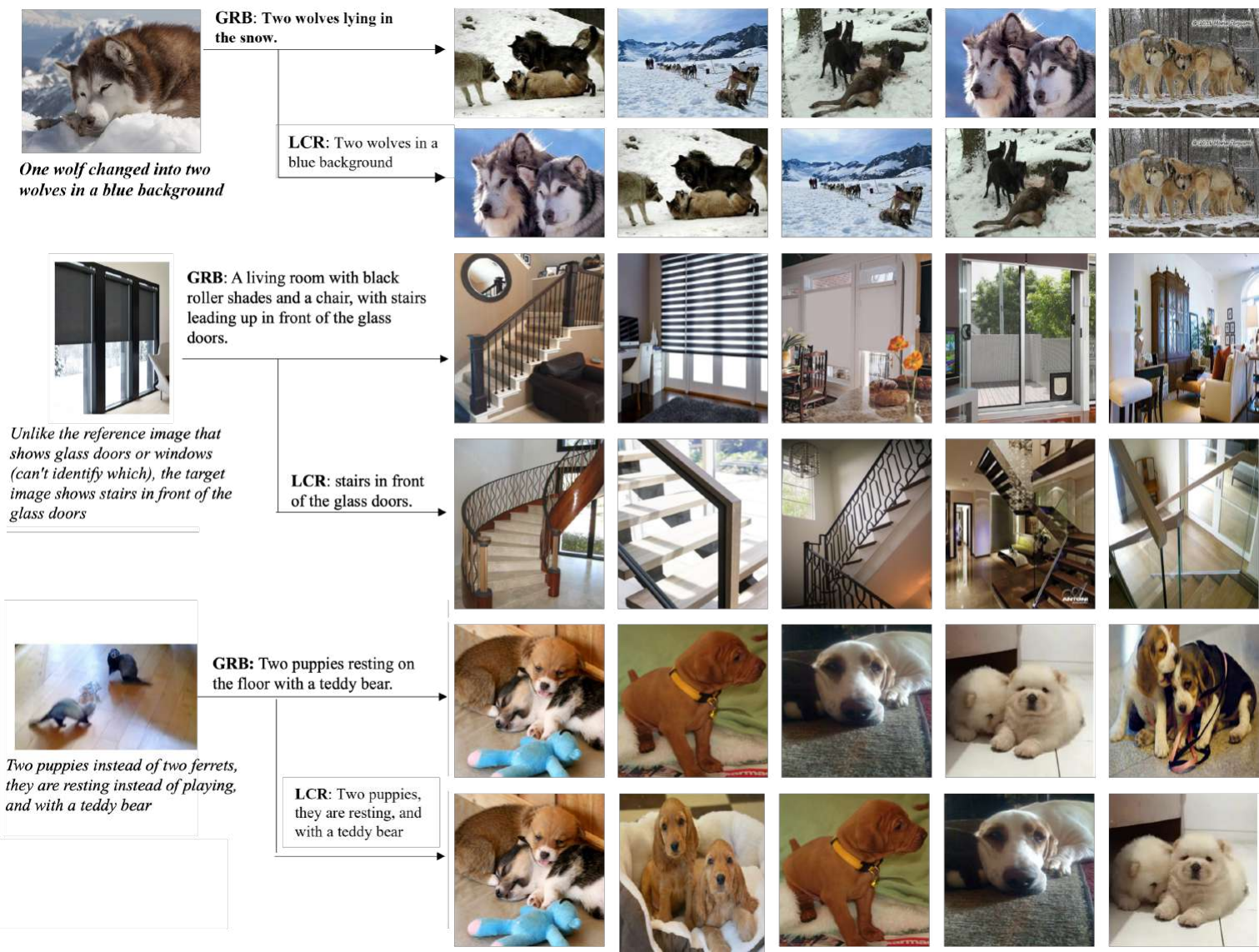}
	\caption{Qualitative results on CIRR test set. The \textbf{GRB} is the global retrieval baseline, which can merge holistic information from reference image caption and modified text. \textbf{LCR} is local concept re-ranking, which can extract discriminative local information to rerank based on the results of the global retrieval baseline. }  \vspace{-0.2cm}
	\label{fig:visres}
\end{figure*}

 \subsection{Implementation Details}
We leverage BLIP2 with OPT-2.7b~\cite{zhang2022opt} as the captioner and its image-text matching module with aligned feature space to retrieve target images. 
For LLM supplying context knowledge, we leverage ChatGPT4-turbo to generate pseudo target captions and local discriminative attributes (concepts). 
For the generated text, we employ a cleaning process to standardize the format and enhance the quality of the text.
We set the scale parameter $\gamma$ to 0.1 for the COCO dataset and 1 for all other datasets. 
To achieve scores corresponding to local concepts, we utilize LLaVA-v1.5-13b and rerank top 50 target images. 
%
All the experiments are conducted on NVIDIA A100 GPUs.












\subsection{Comparison with the SOTAs}
\textbf{Competitors.} We compare the proposed method with several state-of-the-art zero-shot composed image retrieval methods, which are classified into three groups:
\begin{itemize}
  \item Trained on text-image pairs: Pic2Word~\cite{saito2023pic2word}, SEARLE~\cite{baldrati2023zero} and PALAVRA~\cite{cohen2022my}. All of them are textual inversion modules based on the foundation model CLIP~\cite{radford2021learning}. 
  They convert each image to a pseudo word token by fine-tuning on text-image pairs and retrieve in the CLIP pre-learned feature space. 
  PALAVRA is aimed for the personalized image retrieval task with ZS-CIR results reported from~\cite{baldrati2023zero}.
  SEARLE~\cite{baldrati2023zero} additionally includes GPT-powered regularization and knowledge distillation.

  \item Trained on task-specific triplets: CASE~\cite{levy2023data} and TransAgg~\cite{liu2023zero}.
  They both propose new task-specific triplets, namely LaSCo for CASE and Laion-CIR for TransAgg.
  Both of them implement specific layers from the existing foundation model BLIP~\cite{li2022blip}.
  \item Training-free methods: We implement three training-free methods for comparison. 
  Text-only-M conducts retrieval directly with the modified text of the triplets.
  Text-only-C performs retrieval directly with text \{reference caption\} + `change to' + \{modified text\}, i.e., the concatenation of the modified text and caption of the reference image, which is achieved from the BLIP2 captioner.
  LLaVA-CIR is achieved by inputting modified text and image, instructing the LLaVA model to output the target caption directly, with the prompt: \textit{`Please provide a brief image caption if change to \{modified text\}'}. 
  All of these three methods calculate the cosine similarity distance in the aligned feature space of BLIP2~\cite{li2023blip}.
\end{itemize}

\begin{table*}[ht]
  \centering
  \caption{\textbf{Quantitative results on FashionIQ validation set.}} 
    \resizebox{1.0\linewidth}{!}{ 
    \begin{tabular}{llcccccccc} 
    \toprule
    \multicolumn{1}{c}{} & \multicolumn{1}{c}{} & \multicolumn{2}{c}{Shirt} & \multicolumn{2}{c}{Dress} &\multicolumn{2}{c}{Toptee} &\multicolumn{2}{c}{Average} \\
    \cmidrule(lr){3-4}
    \cmidrule(lr){5-6}
    \cmidrule(lr){7-8}
    \cmidrule(lr){9-10}
    \multicolumn{1}{l}{Supervision} & \multicolumn{1}{l}{Method} & R$@10$ & R$@50$ & R$@10$ & R$@50$ & R$@10$ & R$@50$ & R$@10$ & R$@50$ \\      
    \midrule
    \multirow{1}{*}{Triplets } & \notcomp{TransAgg-Combined}~\cite{liu2023zero} & - & - & - & - & -  & - & \notcomp{34.36} & \notcomp{55.13} \\ 
    \midrule
    \multirow{3}{*}{Text-image pairs} & PALAVRA~\cite{cohen2022my} & 21.49 & 37.05 & 17.25 & 35.94 & 20.55 & 38.76 & 19.76 & 37.25 \\

     &Pic2Word~\cite{saito2023pic2word} & 26.20 & 43.60 & 20.00 & 40.20 & 27.90 & 47.40 & 24.70 & 43.70 \\
     &SEARLE-XL-OTI~\cite{baldrati2023zero} & 30.37 & 47.49 & 21.57 & 44.47 & 30.90 & 51.76 & 27.61 & 47.90 \\
    \midrule
    \multirow{4}{*}{Training-free} &Text-only-M & 25.76 & 41.61&  18.52&35.94 & 27.59  & 46.91& 23.93 & 41.44 \\ 
    &Text-only-C & 21.64 & 39.60 & 11.80  & 27.42 & 20.14 & 38.25 & 17.85 & 35.07 \\ 
    &LLaVA-CIR & 9.32 & 19.33 & 6.15  & 14.72 & 9.48 & 19.22 & 8.32 & 17.75 \\ 

    \cmidrule(lr){2-10}
    &\baseline & 34.54 & \textbf{55.15} & 24.14 & \textbf{45.56} & \textbf{33.55} & \textbf{53.60} & 30.74 & \textbf{51.44} \\ %
    & \baseline + \rerank & \textbf{35.38} & \textbf{55.15}  & \textbf{24.84} &  \textbf{45.56} & 33.10 & \textbf{53.60} & \textbf{31.11}  & \textbf{51.44} \\ 

    \bottomrule
    \end{tabular}}
\vspace{-0.5cm}
    \label{tab:fashioniq_val_result}
  \end{table*}

\subsection{Main Results}


\noindent\textbf{CIRR.} 
\label{sec:cirr}
The experimental results for the CIRR test set are presented in Table~\ref{tab:cirr_test_exp} and visualization results are shown in Figure~\ref{fig:visres}. 
As observed, our global retrieval baseline (\baseline) achieves comparable performance with state-of-the-art ZS-CIR methods SEARLE~\cite{baldrati2023zero} and Pic2Word~\cite{saito2023pic2word}.
Notably, both of them are trained on text-image pairs, while our method is training-free for saving computational resources.  
Furthermore, with local concepts re-ranking (\rerank) incorporating discriminative information, our method gains a significant improvement of 8.63\% in Recall@1. 
%
%
In comparison with three other training-free methods, our method \baseline+~\rerank has an average improvement of 11.32\% over Text-only-M, 16.99\% over Text-only-C, and 24.80\% over LLaVA-CIR respectively.
To process modification reasoning, LLaVA-CIR leverages a vision-language model and Text-only-C relies on simple combination.
%
However, it is evident that existing text-image alignment space (BLIP2) cannot handle instructional reasoning effectively, and relying on vision-language model (LLaVA) only leads to suboptimal results.  
%
%
%
Our \baseline addresses modification reasoning by transforming the instructions into descriptive pseudo-target captions using a prompt-enhanced pure language model, resulting in improved performance.  
Notably, our proposed baseline model can outperform existing training-needed methods~\cite{baldrati2023zero,saito2023pic2word} by an average improvement of 4.74\% over SEARLE and 5.57\% over Pic2word.
%
%

\noindent\textbf{FashionIQ.} We present the results on the FashionIQ validation set in Table~\ref{tab:fashioniq_val_result}.
We can observe that our \baseline baseline alone achieves better performance than the existing state-of-the-art ZS-CIR SEARLE, with an average improvement of 6.81\% in Recall@5 and 3.54\% in Recall@10.
FashionIQ is a fine-grained dataset with similar clean backgrounds and fine-grained variations.
We can observe that modified text alone provides strong discriminative information. 
With the increase of reference image information, the model gets distracted and leads to lower performance from Text-only-C to LLaVA-CIR.

%



\begin{wraptable}{r}{0.6\textwidth} 
  \centering
  \small
  \vspace{-1.1cm}
  \caption{\textbf{Quantitative results on CIRCO test set.} Best scores are in bold. } 
     \resizebox{1.0\linewidth}{!}{ 

  \begin{tabular}{llcccc} 
   \toprule
   \multicolumn{1}{c}{} & \multicolumn{1}{c}{} & \multicolumn{4}{c}{mAP$@k$}  \\
   \cmidrule(lr){3-6}

   \multicolumn{1}{l}{Supervision} & \multicolumn{1}{l}{Method} & $k=5$ & $k=10$ & $k=25$ & $k=50$ \\ 
   \midrule
  \multirow{3}{*}{Text-image pairs} & PALAVRA~\cite{cohen2022my} & 4.61 & 5.32 & 6.33 & 6.80 \\
  &Pic2Word~\cite{saito2023pic2word}  & 8.72 & 9.51 & 10.64 & 11.29 \\
  &SEARLE~\cite{baldrati2023zero} & 11.68 & 12.73 & 14.33 & 15.12\\
%
    \midrule[.02em]
  \multirow{5}{*}{Training-free} 
  &Text-only-M   & 4.40 & 4.81 & 5.59 & 5.99 \\   
  &Text-only-C   & 15.28 & 16.45 & 18.48 & 19.39 \\ 
  &LLaVA-CIR   & 12.13  & 12.88 & 14.45 & 15.2 \\ 
  \cmidrule(lr){2-6}

  & \baseline  &  23.76 & 25.31 & 28.19 & 29.17 \\ 

   & \baseline + \rerank  &  \textbf{26.52} & \textbf{28.25} & \textbf{31.23} & \textbf{31.99} \\ 
  


\bottomrule
   \end{tabular}}
   \vspace{-0.5cm}
   \label{tab:circo_test_exp}
 \end{wraptable}
\noindent\textbf{CIRCO.} As shown in Table~\ref{tab:circo_test_exp}, we present the performance on the CIRCO test set. 
As an open domain dataset with multiple ground truths from a much larger gallery set, CIRCO imposes higher requirements of fine-grained discriminative ability.
Both of our methods show obvious improvements over existing methods.
Specifically, our baseline with global-level retrieval \baseline outperforms SEARLE by 12.08\% in mAP@5. 
Additionally, re-ranking with local concepts \rerank further yields 2.89\% average improvement. 
The substantial enhancement observed in CIRCO underscores the superiority of addressing reasoning in text over aligning the two modalities.

\begin{figure}[t]
  \centering
  \includegraphics[width=0.98\textwidth]{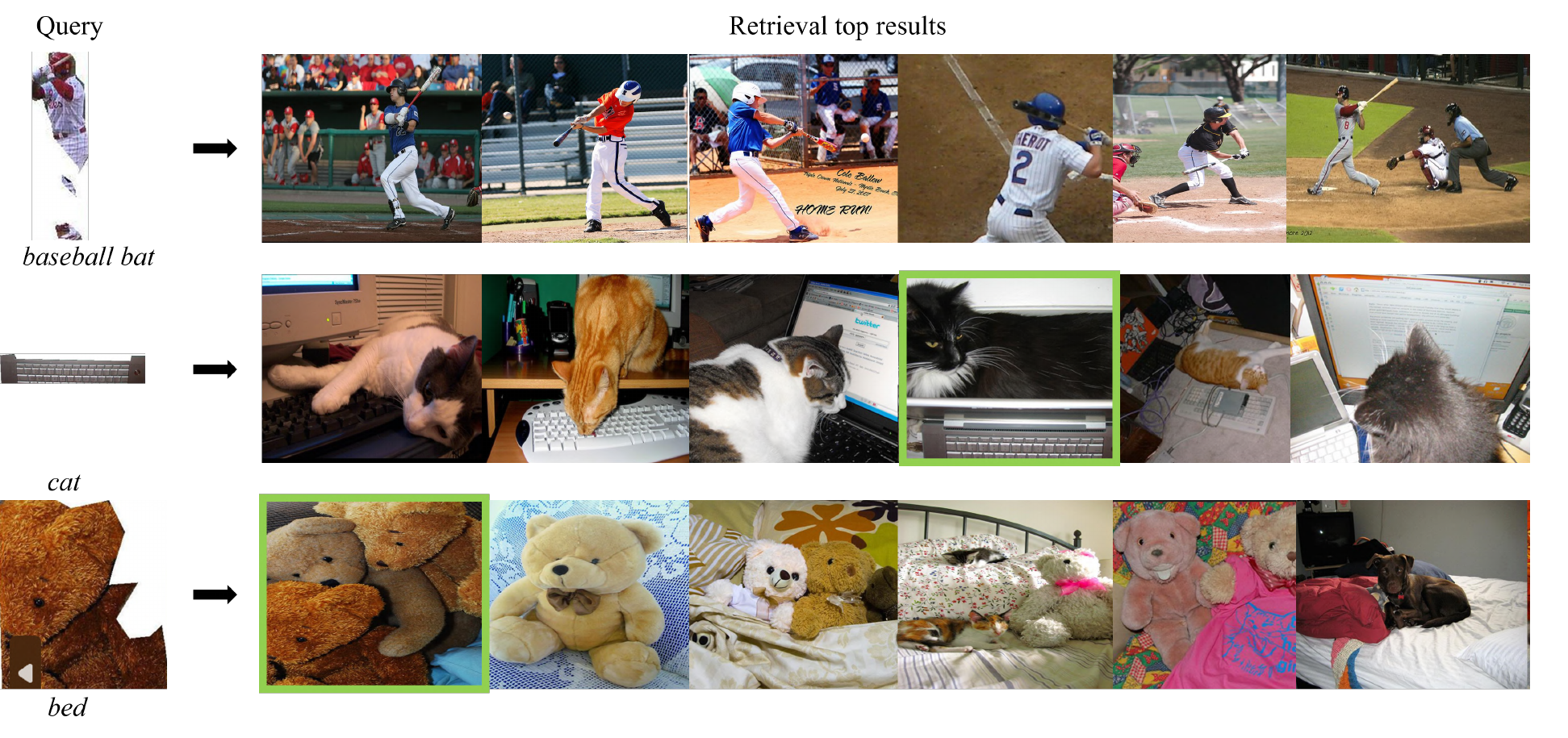}
  \caption{Qualitative results on COCO for object composition.} \vspace{-0.4cm}
  \label{fig:coco}
\end{figure}

\noindent\textbf{COCO.} Table~\ref{tab:coco} and Figure~\ref{fig:coco} present the quantitative and qualitative performance on COCO~\cite{lin2014microsoft} for object composition following~\cite{saito2023pic2word}, respectively.  
Note that the modified text is simplified as the name of classes, thus our \baseline is
\begin{wraptable}{r}{0.5\textwidth} 
  \vspace{-1.0cm}
    \caption{Quantitative result on COCO.}\label{tab:coco}
  \resizebox{1.0\linewidth}{!}{ 
\begin{tabular}{c c c c c}
  \toprule
   Supervision & Method & R@1 &R@5 & R@10 \\    
  \hline
  Text-image pairs & Pic2word & 11.5& 24.8& 33.4 \\
 \hline
    \multirow{2}{*}{Training-free} & GRB & 15.6& 30.7& 38.8\\
   & GRB+LCR& \textbf{29.2}& \textbf{55.6}& \textbf{69.8}\\
  \bottomrule
  \end{tabular}} \vspace{-1cm}
\end{wraptable}
equivalent to Text-only-C in this scenario. 
With \rerank, our method achieves a significant 17.7\% improvement in R@1 over the existing state-of-the-art method Pic2Word.


\vspace{-0.3cm}
\subsection{Ablation Studies}

\noindent\textbf{Evaluating Variants of Captioner and LLMs.}
In this section, as shown in Table~\ref{tab:ablation_LLM}, we evaluate various captioners and large language models for the global retrieval baseline.
We implement instructBLIP~\cite{li-etal-2023-lavis} to generate a detailed caption by employing the prompt `Describe the content of the image in detail', contrasting with BLIP2, which provides a concise caption.
Furthermore, different versions of ChatGPT are deployed to explore the impact of large language models on the reasoning process.
We observe that instructBLIP provides detailed reference information but can mislead the model.
Meanwhile, the large language model GPT4 outperforms GPT3.5, indicating enhanced reasoning capabilities.

\begin{figure}[tbp]
  \centering
  \begin{subfigure}[b]{0.33\textwidth}
      \centering
      \includegraphics[width=\textwidth]{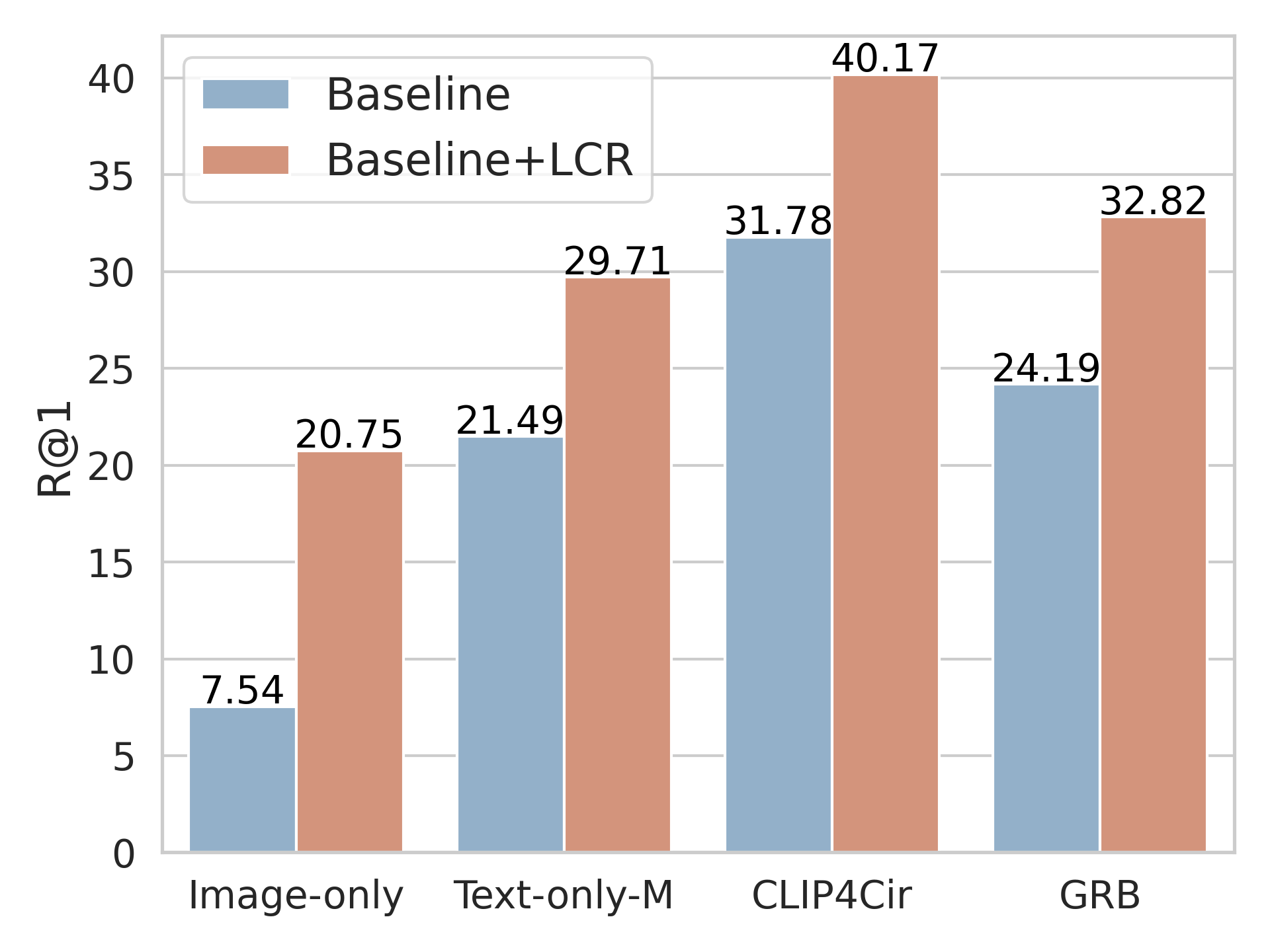} 
      \caption{Baseline Variations} 
      \label{fig:sub1} 
  \end{subfigure}
  \hfill
  \begin{subfigure}[b]{0.33\textwidth}
      \centering
      \includegraphics[width=\textwidth]{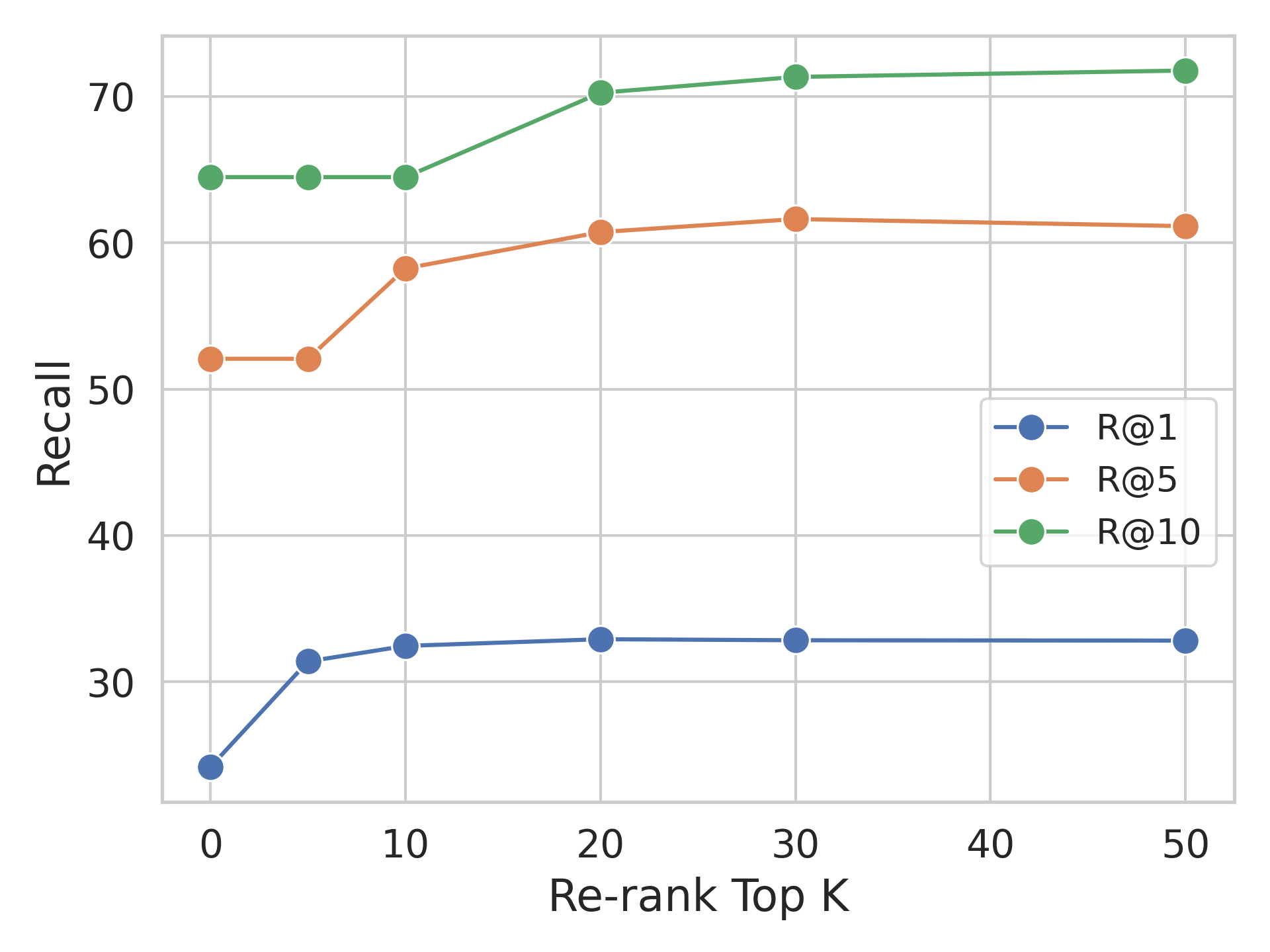} 
      \caption{Re-rank Top $K$ Variations} 
      \label{fig:sub2} 
  \end{subfigure}
  \hfill
  \begin{subfigure}[b]{0.32\textwidth}
      \centering
      \includegraphics[width=\textwidth]{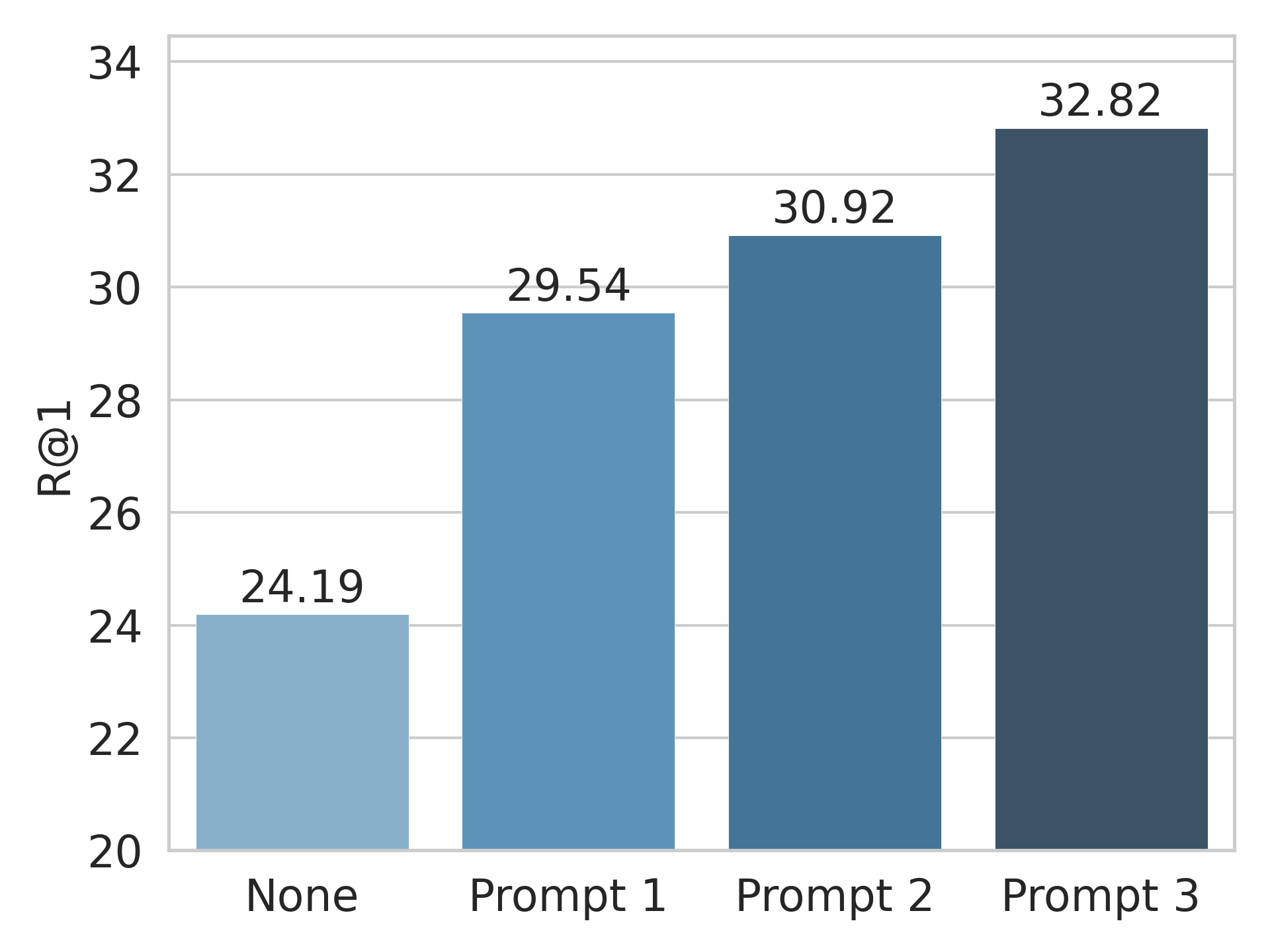} 
      \caption{Prompt Variations} 
      \label{fig:sub3} 
  \end{subfigure}

  \caption{Ablation study on CIRR test set.} 
  \vspace{-0.2cm}
  \label{fig:ablation} 
\end{figure}

\noindent\textbf{Evaluating Variants of Baselines.} Figure~\ref{fig:sub1} compares LCR with additional baselines including training-free methods Image-only, Text-only-M, and a triplet-supervised method CLIP4Cir~\cite{baldrati2022effective}. 
Our LCR shows consistent and significant improvement across both training-free and training-based baselines.

\noindent\textbf{Evaluating Variants of Re-rank Top $K$.} Figure~\ref{fig:sub2} compares LCR with various top $K$.
We observed that the improvement is significant initially and stabilizes with larger $K$, indicating that a small effort (re-rank top 5) in the re-ranking process can yield noticeable improvements (7.2\% in R@1).

\noindent\textbf{Evaluating Variants of Prompts $P_l$.} Figure~\ref{fig:sub3} evaluates the sensitivity of our model to prompts $P_l$.
%
Similar to the default prompt denoted as $P_{l}$ in Figure~\ref{fig:overview}, Prompt1 is without examplars and instructs the model to return positive and negative concepts by noun phrases. Prompt2 is with additional exmplars and returns the output also by noun phrase. Prompt 3, our default prompt, is with additional examplars and instructs the model to return sentences.
%
%
%
%
%
We observe that both detailed information and additional examplars can bring improvement.
%



\begin{table}[!t]
  \centering
  \caption{\textbf{Ablation study on CIRR and CIRCO test set.} } \vspace{-0.2cm}
  \small
  \begin{tabular}{llcccccccc} 
  \toprule
\multicolumn{1}{c}{} & \multicolumn{1}{c}{} & \multicolumn{4}{c}{CIRR Recall} & \multicolumn{4}{c}{CIRCO mAP@$k$}\\ 
  \cmidrule(lr){3-6}
  \cmidrule{7-10}
   Captioner & LLM & R$@1$ & R$@5$ & R$@10$ &  R$@50$ & $k=5$ & $k=10$ & $k=25$ &  $k=50$  \\ \midrule


   InstructBLIP   & GPT3.5 & 20.75 & 45.54 & 58.10  & 79.42 &  11.52 &12.36& 14.19& 14.92  \\
   InstructBLIP   & GPT4 & 22.60 & 49.88 & 62.55 & 84.99 & 16.46 & 17.57& 19.68&  20.56 \\
   BLIP2   & GPT3.5 & 21.49 & 45.78 & 57.11 & 81.33  & 21.32 & 22.81& 25.25& 26.27 \\
   BLIP2 & GPT4 & \textbf{24.19} & \textbf{52.07} & \textbf{64.48} &  \textbf{85.28} & \textbf{23.76}& \textbf{25.31} &\textbf{28.19}  &\textbf{29.17}  \\
  \bottomrule
  \end{tabular} 

  \label{tab:ablation_LLM}
\end{table}

\noindent\textbf{Limitations and Future Work.}
From the experimental results, it becomes evident that a careful construction of query information is imperative, especially to determine the requirement that must be satisfied. 
However, the requirement supplied by the reference image is ambiguous.
In this study, we address this by re-ranking on the local concepts that must be satisfied. 
The conditional selection of reference image information remains an avenue deserving further investigation.

\section{Conclusion}
This paper presents an innovative training-free method for zero-shot composed image retrieval, incorporating re-ranking strategy informed by discriminative local concepts.
Global retrieval baseline converts the instruction-image query into a descriptive caption to leverage existing visual-textual alignment space. 
The second stage, local concept re-ranking, solves the problem of ambiguous requirements from the reference image by re-ranking on precise fine-grained concepts. 
%
%
The proposed method demonstrates comparable performances to that of existing training based zero-shot composed image retrieval methods and outperforms significantly all other training-free methods.



%
%
\bibliographystyle{splncs04}
\bibliography{main}
\end{document}